# Visual Query Answering by Entity-Attribute Graph Matching and Reasoning

Peixi Xiong[1]*, Huayi Zhan[1]*, Xin Wang[2]*,
Baivab Sinha[3], Ying Wu[1]
[1]Northwestern University,[2] Southwest Jiaotong University,
[3]Sichuan Changhong Electric Co. Lt
`peixixiong2018, huayi.zhan, yingwu@u.northwestern.edu,`
`xinwang@swjtu.cn, baivabsinha@changhong.com`

## Abstract

*Visual Query Answering (VQA) is of great significance in offering people convenience: one can raise a question for details of objects, or high-level understanding about the scene, over an image. This paper proposes a novel method to address the VQA problem. In contrast to prior works, our method that targets single scene VQA, replies on graph-based techniques and involves reasoning. In a nutshell, our approach is centered on three graphs. The first graph, referred to as inference graph $G_I$, is constructed via learning over labeled data. The other two graphs, referred to as query graph $Q$ and entity-attribute graph $G_{EA}$, are generated from natural language query $Q_{nl}$ and image Img, that are issued from users, respectively. As $G_{EA}$ often does not take sufficient information to answer $Q$, we develop techniques to infer missing information of $G_{EA}$ with $G_I$. Based on $G_{EA}$ and $Q$, we provide techniques to find matches of $Q$ in $G_{EA}$, as the answer of $Q_{nl}$ in Img. Unlike commonly used VQA methods that are based on end-to-end neural networks, our graph-based method shows well-designed reasoning capability, and thus is highly interpretable. We also create a dataset on soccer match (Soccer-VQA) with rich annotations. The experimental results show that our approach outperforms the state-of-the-art method and has high potential for future investigation.*

## 1. Introduction

In recent years, visual query answering (VQA) has received significant attention [21, 25, 9] as it involves multi-disciplinary research, *e.g.* natural language understanding, visual information retrieving and multi-modal reasoning. The task of VQA is to find an answer to a query $Q_{nl}$ based on the content of an image. There are a variety of applications of VQA, *e.g.* surveillance video understanding, visual commentator robot, *etc.* Solving VQA problems usually requires high level reasoning from the content of an image.

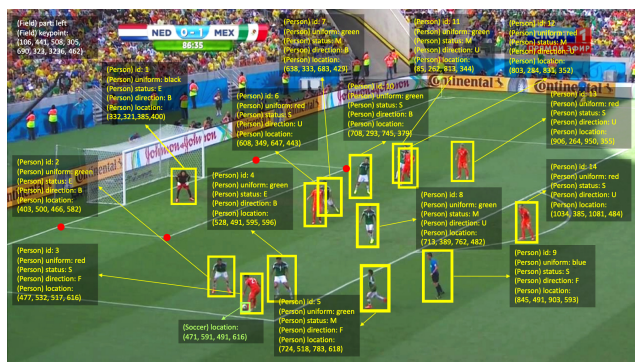

Figure 1: The image is about soccer match, where each person object is associated with attributes: id, uniform color, status (Standing, Moving, Expansion), direction (Backing, Facing, N/A), as well as location, and the soccer object is attributed with location.

Ever since the VQA problem was first brought up by [20], end-to-end neural network has become the dominant approach in the community. The basic idea is to extract representations of image and text by convolutional neural network and recurrent neural network respectively, and to combine these two representations to form a joint embedding which is then fed to a classifier to infer the answer [36]. The NN-based approach tries to learn the correlations of text query with input image (and implicitly with the expected answer) in a joint semantic space. A major improvement to the basic method is to add attention mechanism [38, 31, 37, 13, 19].

Despite the dominance in VQA literature, neural network approach has a few weaknesses, which greatly hinders its further development. First of all, deep neural network works as "black boxes", hence it is very hard to identify the causal relations between model design and system performance. Secondly, but more importantly, there is no evidence to support the hypothesis that neural network has the capability of reasoning in solving VQA problems. On the contrary, recent work [11] has shown that it is possible to do quite well on many VQA problems by simply memorizing statistics about query / answer pairs. To overcome these weaknesses, a more feasible method with viable reasoning capability is highly needed. Consider the problem of natural language query answering (NLQ), which is analogous to VQA problem but without image input, the state-of-the-art approach to NLQ problem prefers to apply graph-based techniques, that represents underlying answers and queries as knowledge graph and query graph, respectively, and find answers with graph pattern matching, rather than relying on conventional neural network based methods. The benefit of the approach lies in that structured representations contain richer information than unstructured ones, and hence is capable to find reliable results. Indeed, the similar technique can also be applied for VQA problem.

**Example 1:** Figure 1 depicts an image about a soccer match, where two teams are distinguished by red and green uniforms, and each object is associated with a set of attributes. A typical query may ask "How many players are there in the image?". Though simple, it is nontrivial to answer the query, as we not only need to identify all the *person* objects, but also have to infer the hidden attribute "role" of each person, *i.e.* reasoning whether the person is a player, or a goalkeeper, or a referee.

To answer the query, one can represent the image with graph structure by identifying objects along with their attributes, and constructing a graph $G_{EA}$, denoted by entity-attribute graph, using objects that are identified.

The benefits of graph representation are twofold: (1) as $G_{EA}$ may not contain sufficient information to answer query, *e.g.* value of attribute "role " may not be identified via visual method, we are allowed to develop techniques to reason missing information that is crucial for the query; and (2) query answering can be evaluated via graph pattern matching due to structured representation of the query. □

This example suggests that we leverage graph-based method to resolve the VQA problem. While to do this, several questions have to be settled. (1) How to represent image and query with graphs? (2) How to infer crucial information when $G_{EA}$ constructed from image is insufficient? (3) How to find answers from graphs with $G_{EA}$?

The contributions of our paper include following aspects:

(1) We produced a data set of 7900 images on soccer match. For each image in the data set, we make a detailed annotation on objects to describe their attributes, *e.g.* color, role, status, location, *etc*. To the best of our knowledge, this is the first data set about soccer match in VQA literature.

(2) We propose approaches to answering visual questions with graph-based techniques. More specifically, we

first construct an entity-attribute graph from a given image; we then train a classifier to infer missing information that are crucial for answering queries; we finally provide methods to answer queries with graph pattern matching.

## 2. Related Work

We categorize related work into following three parts.

*Visual query answering*. Current VQA approaches are mainly based on deep neural works. [38] introduces a spatial attention mechanism similar to the model for image captioning. Instead of computing the attention vector iteratively, [31] obtains a global spatial attention weights vector which is then used to generate a new image embedding. [37] proposed to model the visual attention as a multivariate distribution over a grid-structured conditional random field on image regions, thus multiple regions can be selected at the same time. This attention mechanism is called structured multivariate attention in [37]. There has been many other improvements to the standard deep learning method, *e.g.* [8] utilized Multimodal Compact Bilinear (MCB) pooling to efficiently and expressively combine multimodal features. Another interesting idea is the implementation of Neural Module Networks [3, 12], which decomposes queries into their linguistic substructures, and uses these structures to dynamically instantiate module networks. [27] proposed to build graph over scene objects and question words. The visual graph is similar to ours, but the query graph differs. Note that the method [27] proposed is still a neural network based method as the structured representations are fed into a recurrent network to form the final embedding and the answer is again inferred by a classifier.

*Visual Objects Processing*. Visual object detection as well as relationship identification are the preliminary tasks for not only VQA but also image captioning [18, 32, 27]. Other works, *e.g.* [33], produce high-level attributes for input images, based on which further processing can be conducted. These prior works show that detecting all visual objects, their attributes and relationships is very vital for resolving VQA problem.

*Graph-based query answering*. Query answering has been extensively studied for graph data. In a nutshell, this work includes two aspects: query understanding, and query evaluation. We next review previous work on two aspects.
(1) Queries expressed with natural languages are very user-friendly, but nontrivial to understand. Typically, they need to be structured before issuing over *e.g.* search engine, knowledge graph, since structured queries are more expressive. There exist a host of works that based on query logs, human interaction and neural network, respectively. [24] leverages query logs to train a classifier, based on which structured queries are generated. [35] propose an approach to generate the structured queries through talking between the data (*i.e.* the knowledge graph) and the user. [34] introduced how to generate a core inferential chain from a query with convolutional neural networks. As we only cope with a set of fixed queries, hence, we defer the topic of query understanding to another paper, and focus primarily on the query evaluation.

(2) To evaluate queries on graphs, a typical method is graph pattern matching. There has been a host of work on graph pattern matching, *e.g.* techniques for finding exact matches [5, 29], inexact matches [39, 28], and evaluating SPARQL queries on RDF data [30]. Our work differs from the prior work in the following: (1) we integrate arithmetical and set operations in the query graph, and (2) we develop technique to infer missing values for query answering.

## 3. New Dataset

In this section, we introduce our dataset as well as typical domain specific questions.

### 3.1. Innovations

Traditional VQA datasets, *e.g.* [4, 11, 26, 10, 16] are of large scale. Though workforce and resource intensive, these datasets are inappropriate for rule learning and reasoning due to characteristics of overbroad domain and insufficient scene meaning. Some other datasets, *e.g.* [14, 20, 2], narrow the domain for better reasoning. However, images in these datasets are very elementary, with simple relationship among objects in the image, as a consequence, they are not very helpful to find interesting rules after reasoning. Compared with theirs, ours has following two main innovations: (1) our dataset is not only domain specific, but also includes images that are pretty content-rich, these together enables us to do reasoning very well; (2) with rules inferred, complex questions, that implicate reasoning, arithmetic operating, *etc.*, can be answered with high accuracy.

### 3.2. Images

**Scale**. A set of 7900 frames were collected from 2016 FIFA World Cup videos, among which, 5900 frames are chosen as training set, 1000 frames for validation and remaining for testing. To ensure validity of testing, we discarded similar frames from the same sequence.

**Annotation**. Annotation of our dataset consists of four main parts, based on the object type: *person*, *field*, *soccer* and *scene* (Table 1). Here, we localize an object by a bounding box and record the minimum and maximum values of four corners. To distinguish each person, we annotate the role he plays, the relative direction between him and the goal, his action and his uniform color, *etc.* To better locate objects, we record whether this image is about the left, right or middle part of the field, along with corresponding four keypoints. To better evaluate the high-level meaning of the image, we also record the scene type of it.

| Object | Attribute | Type | Descriptions |
|--------|-----------|------|--------------|
| Person | id | obvious | An index for each person in the field. |
| | role | hidden | *e.g.* player, goalkeeper, referee |
| | uniform | obvious | The uniform color of this person. *e.g.* red, blue |
| | location | obvious | The coordinates of the bounding box. *e.g.* (xmin, ymin, xmax, ymax) |
| | direction | obvious | The direction between this person and the goal. *e.g.* backing, facing, n/a |
| | status | obvious | The current action of this person. *e.g.* standing, moving, expansion |
| | defending | hidden | Whether this person is defending others. *e.g.* yes, no |
| Field | part | obvious | Which part of the field is this image about. *e.g.* left, right, middle |
| | keypoint | obvious | Record locations for four corners of penalty area; Or, the lengths of center circle's major and minor axis and its center. |
| Soccer | location | obvious | The coordinates of the bounding box. *e.g.* (xmin, ymin, xmax, ymax) |
| Scene | type | hidden | *e.g.* normal scene, free kick, kick off, corner kick, penalty kick |

Table 1: Visual objects and their attributes.

### 3.3. Questions

Our questions, which are of 7 types, involve counting, detection, role identification and understanding of the scene. To better evaluate performance of the model, we categorized the questions into three levels, easy, medium and hard (Table 2). They are decided by the number of vision tasks needed during the process, and the level of knowledge graph usage for reasoning. For the answer part, we asked 5 people to manually answer the questions, so the answers may vary in format.

| Id | Question | Difficulty |
|----|----------|------------|
| $Q_{nl_1}$ | Who is holding the soccer? | Easy |
| $Q_{nl_2}$ | What is the uniform color of the referee? | Easy |
| $Q_{nl_3}$ | Is there any referee in the image? | Easy |
| $Q_{nl_4}$ | Which team does the goalkeeper belong to? | Medium |
| $Q_{nl_5}$ | Who is the defending team? | Medium |
| $Q_{nl_6}$ | Which part of the field are the players being now? | Hard |
| $Q_{nl_7}$ | How many players are there in the image? | Hard |

Table 2: A set of questions

**Evaluation Criteria**. The accuracy is calculated by checking if the predicted answer is the same as any of human-provided answers. In our experiments, to eliminate errors that are caused by machines' indistinguishability on variance of the ground truth answers, we asked 20 people with different gender and age to manually check if the question is correctly answered.

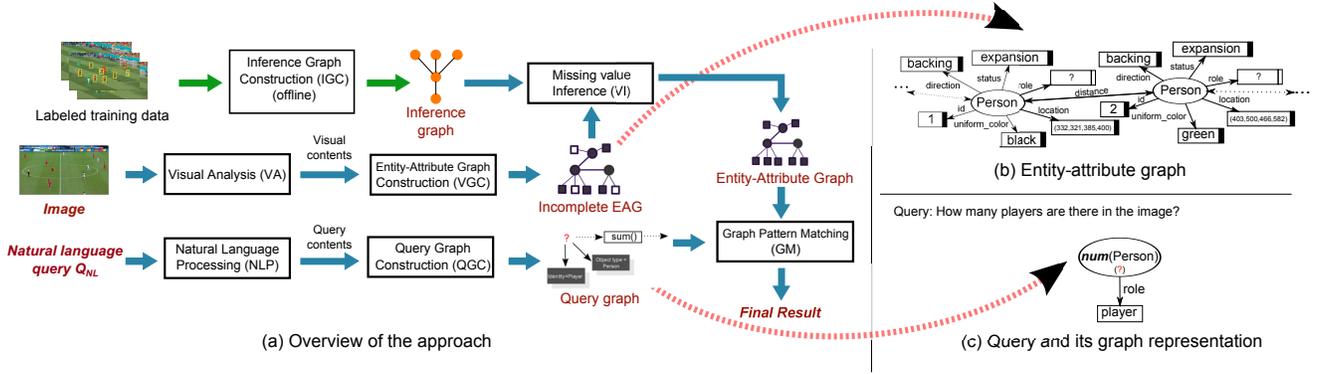

Figure 2: Overview of our approach, Entity Attribute Graph and Queries

# 4. Our Approach

In this section, we introduce our approach with details.

## 4.1. Representation

Below, we first review a few concepts.

### 4.1.1 Entity-Attribute Graph

We start with notions of entities, attributes, relations and entity-attribute graphs.

**Entities, Attributes & Relations.** Entities are typically defined as objects or concepts that exist in the real world, *e.g.* people, soccer etc. An entity often carries multiple attributes, that describe characteristics of the entity, *e.g.* uniform color, person role. Among entities, there may exist various relationships, *e.g.* friendship, showing the correlation of entity pairs.

**Entity-Attribute Graphs.** Assume a set $\mathcal{E}$ of entities, a set $\mathcal{D}$ of values, a set $\mathcal{P}$ of predicates indicating attributes of entities and a set $\Theta$ of types. Each entity $e$ in $\mathcal{E}$ has a *unique ID* and a *type* in $\Theta$.

An *entity-attribute* graph, denoted as EAG, is a set of triples $t = (s, p, o)$, where *subject* $s$ is an entity in $\mathcal{E}$, $p$ is a *predicate* in $\mathcal{P}$, and *object* $o$ is either an entity in $\mathcal{E}$ or a value $d$ in $\mathcal{D}$. It can be represented as a directed edge-labeled graph $G_{EA} = (V, E)$, such that (a) $V$ is the set of nodes consisting of $s$ and $o$ for each triple $t = (s, p, o)$; and (b) there is an edge in $E$ from $s$ to $o$ labeled by $p$ for each triple $t = (s, p, o)$.

We consider two types of equality:

(a) *node identity* on $\mathcal{E}$: $e_1 \Leftrightarrow e_2$ if entities $e_1$ and $e_2$ have the same ID, *i.e.* they refer to the same entity; and

(b) *value equality* on $\mathcal{D}$: $d_1 = d_2$ if they are the same value.

In $G_{EA}$, $e_1$ and $e_2$ are represented as the same node if $e_1 \Leftrightarrow e_2$; similarly for values $d_1$ and $d_2$ if $d_1 = d_2$.

**Example 2:** Figure 2 (b) shows a sample EAG, where each rounded (resp. square) node represents an entity (resp. attribute), each directed edge labeled by $p$ from an entity node $v_e$ to a value node $v_a$ denotes that $v_e$ has a $p$ attribute with value $v_a$, and each object pair is connected with bidirectional arrow due to mutual relationship, *e.g.* distance. □

**Image Representation.** An image can be represented as an EAG with detected objects and obvious attributes. This can be achieved via a few visual tasks. While EAG generated directly after image processing is often incomplete, *i.e.* it may miss some crucial information to answer certain queries. We hence refer to *entity-attribute graphs* with incomplete information as *incomplete entity-attribute graphs*, and associate nodes with white rectangles, to indicate the missing value of an entity or attribute in EAG. Figure 2(b) is an *incomplete entity-attribute graph*, in which square nodes representing person roles are associated with white rectangle.

As queries issued with natural languages are often translated into graph structures for the purpose of evaluation, to answer structured queries, it would be beneficial to construct an EAG from an image so that existing techniques can be directly applied for query answering.

### 4.1.2 Query Representation

It is recognized that querying graph data with keywords from $Q_{nl}$ may not well capture users query intention [24]. Instead, a structured query with "query focus" is favored. In light of this, we next introduce the notion of query graphs.

**Query Graphs.** A query graph $Q(u_o)$ is a set of triples $(s_Q, p_Q, o_Q)$, where $s_Q$ is either a variable $z$ or a function $f(z)$ taking $z$ as parameter, $o_Q$ is one of a value $d$ or $z$ or $f(z)$, and $p_Q$ is a predicate in $\mathcal{P}$. Here function $f(z)$ is defined by users, and variable $z$ has one of three forms: (a) *entity variable* $y$, to map to an entity, (b) *value variable* $y*$, to map to a value, and (c) *wildcard* $\_y$, to map to an entity. Here $s_Q$ can be either $y$ or $\_y$, while $o_Q$ can be $y$, $y*$ or $\_y$. Entity variables and wildcard carry a *type*, denoting the type of entities they represent.

A query graph can also be represented as a graph such that two variables are represented as the same node if they have the same name of $y$, $y*$ or $-y$; similarly for functions $f(z)$ and values $d$. We assume *w.l.o.g.* that $Q(x)$ is connected, *i.e.* there exists an undirected path between $u_o$ and each node in $Q(u_o)$. In particular, $u_o$ is a designated node in $Q(u_o)$, denoting the query focus and labeled "?". Take Fig. 2(c) as example. It depicts a query graph that is generated from query "*How many players are there in the image?*". Note that the "query focus" $u_o$ carries a function num () that calculates the total number of *person* entities with *role* "player".

**Remark.** In this paper, we do not cope with arbitrary $Q_{nl}$, and only handle a set of fixed queries (Table 2). In light of this, we do not provide techniques to structure $Q_{nl}$. We refer interested readers to references, *e.g.* [24, 35, 34], for more details about the task.

#### 4.1.3 Graph Pattern Matching

We introduce the notion of *valuation*, followed by graph pattern matching problem (GPM).

**Valuation.** A valuation of $Q(u_o)$ in a set $S$ of triples is a mapping $\nu$ from $Q(u_o)$ to $S$ that preserves values in $\mathcal{D}$ and predicates in $\mathcal{P}$, and maps variables $y$ and $-y$ to entities of *the same type*. More specifically, for each triple $(s_Q, p_Q, o_Q)$ in $Q(u_o)$, there exists $(s, p, o)$ in $S$, written as $(s_Q, p_Q, o_Q) \mapsto_\nu (s, p, o)$ or simply $(s_Q, p_Q, o_Q) \mapsto (s, p, o)$, where

(a) $\nu(s_Q) = s, p = p_Q, \nu(o_Q) = o$;

(b) $o$ is an entity if $o_Q$ is a variable or $-y$; it is a value if $o_Q$ is $y*$, and $o = d$ if $o_Q$ is a value $d$; and

(c) entities $s$ and $s_Q$ have the same type; similarly for entities $o$ and $o_Q$ if $o_Q$ is $y$ or $-y$.

We say that $\nu$ is a *bijection* if $\nu$ is one-to-one and onto.

**Graph Pattern Matching.** [5]. Consider an EAG $G_{EA} = (V, E)$ and a query graph $Q(u_o)=(V_Q, E_Q, u_o)$. We say that $G_{EA}$ matches $Q(u_o)$ at $e$ if there exist a set $S$ of triples in $G_{EA}$ and a valuation $\nu$ of $Q(u_o)$ in $S$ such that $\nu(x) = e$, and $\nu$ is a bijection between $Q(u_o)$ and $S$. We refer to $S$ as a match of $Q(u_o)$ in $G_{EA}$ at $e$ under $\nu$. Intuitively, $\nu$ is an isomorphism from $Q(u_o)$ to $S$ when $Q(u_o)$ and $S$ are depicted as graphs. That is, we adopt subgraph isomorphism for the semantics of graph pattern matching.

### 4.2. VQA Modeling

We propose a comprehensive approach as modeling of the VQA problem.

The inference graph $G_I$ is used to infer missing values of an *incomplete* EAG. and constructed by module IGC over training data. As is query-independent, $G_I$ is constructed offline, which warrants the efficiency of our approach. As the other part of input, natural language query $Q_{nl}$ needs to be structured for query evaluation. To this end, $Q_{NL}$ is first parsed via our NLP module, and then structured by module QGC. After $Q(u_o)$ and $G_{EA}$ are generated, our approach employs module GM for matching computation, and returns final result.

As some modules employ existing techniques, to emphasize our novelty, we will elaborate modules VA and VGA in Section 4.3, modules IGC and VI in Section 4.4, and module GM in Section 4.5 with more details.

### 4.3. EAG Generation from Images

We next introduce how an EAG is constructed by illustrating functions of modules VA and VGA.

#### 4.3.1 Visual Processing

Inspired by [32, 17, 6], module VA conducts a few visual tasks to detect the objects and figure out their attributes. Influenced by queries given in Table 2, for each image img, module VA only recognizes four types of objects, *i.e.person*, *field*, *soccer* and *scene*, as shown in Table 1.

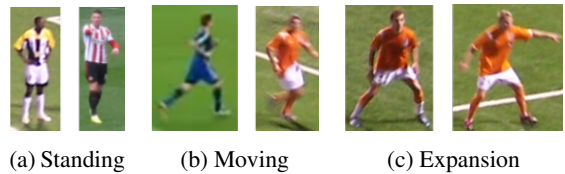

(a) Standing    (b) Moving    (c) Expansion

Figure 3: Person status.

Many obvious attributes of *person* object can be obtained by simple vision tasks. For instance, attributes "location", "direction", and "status" can be figured out by object detection, followed by skeleton detection in the object regions and appropriate classifying for skeleton patterns. As shown in Fig. 3, we categorize three types of *person* "status", *i.e.* standing, moving and expansion, where the last one is distinguished from the first two by the pattern of object's knees and the space he occupied.

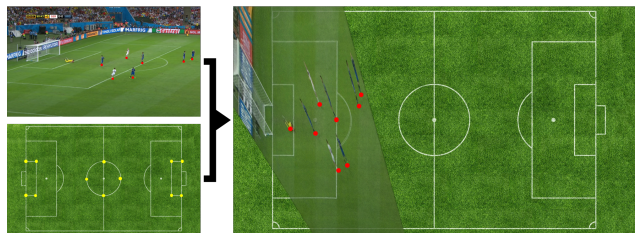

Figure 4: Image registration to standard field.

Obvious attributes of *field* object can be detected as follows. Attribute "part" is distinguished via simple image classifier. Attribute "keypoint" can be identified by edge and circle detection. With "keypoint", we register the image into our standard field (Figure 4), then all local coordinates (locations of *person* and *soccer*) in the image can be transformed into a global coordinates of the bird's-eye view standard soccer field.

After processing, VA module outputs a set of identified objects and their obvious attributes for EAG construction.

#### 4.3.2 EAG **Construction**

Module VGA is responsible for EAG construction. Given output of module VA over image, VGA conducts the following: (1) constructing an empty entity-attribute graph $G_{EA}$; (2) treating objects and attribute values as *subject* and *object*, respectively, and creating nodes corresponding to each object and attribute value in $G_{EA}$; (3) connecting node $v_e$ to node $v_a$ with edge labeled by $p$, to indicate that entity $e$ has an attribute $p$ with value $a$ (nodes $v_e$ and $v_a$ correspond to $e$ and $a$, respectively), for each entity and its obvious attribute; and (4) linking node pair $(v_{e_1}, v_{e_2})$, with bidirectional edge labeled with distance between entity $e_1$ and $e_2$. Note that, VGA also connects entity node $v_e$ to a value node $v_b$ taking blank value with edge labeled by $p'$, if $p'$ is a hidden attribute, and the value of attribute $p'$ can not be identified by module VA.

### 4.4. EAG-based Reasoning

An *incomplete* EAG is often not able to provide query answers due to missing values of some hidden attributes. This motivates us to develop methods to infer values of hidden attributes. Below, we present modules IGC and IM, which are responsible for inference graph construction and missing value inference, respectively.

In our model, the inference graph is constructed using the Bayesian network. Essentially, Bayesian network is a kind of directed acyclic graph model, of which the parameters can be explicitly represented by the nodes (*i.e.*, random variables). Additionally, the parameters can be endowed with distributions (*i.e.*, priors). Using Bayesian network as inference graph leads to the resulting structure being very concise.

#### 4.4.1 **Inference Graph**

As mentioned above, the inference graph is constructed using Bayesian network. A typical Bayesian network consists of decision and utility nodes [22]. We follow the descriptive notations used in [15] to facilitate our problem. Defined by $\mathcal{D} = \{\mathbf{x}^{(i)}\}_{i=1}^N$ the set of $N$ instances, each instance $\mathbf{x}^{(i)} = [x_1^{(i)}, \cdots, x_n^{(i)}]$ is the observation over $n$ random variables: $x_1 \sim X_1, \cdots, x_n \sim X_n$. Under this assumption, a Bayesian network can be formally described by $\mathfrak{B} = <\mathcal{G}, \Theta_{\mathcal{G}}>$, where $\mathcal{G}$ is a directed acyclic graph and $\Theta_{\mathcal{G}}$ the set of parameters that can maximize the likelihood [7, 23]. The $i$-th node in $\mathcal{G}$ corresponds to a random variable $X_i$, and an edge between two connected nodes indicates the direct dependency. The symbol of $\Theta_{\mathcal{G}}$ is a parametric set that uses to quantify the dependencies within $\mathcal{G}$. Specifically, the parameters set of the $i$-th node associated with an observation $x_i$ in $\Theta_{\mathcal{G}}$ can be denoted by $\theta_{x_i | \Pi_i(\mathbf{x})}$, where $\Pi_i(\mathbf{x})$ is a function which takes $\mathbf{x}$ as input, and outputs the values of attributes whose child is $i$. Note here that $x_i$ is a possible value of $X_i$. For notational simplicity, the notation of $\theta_{x_i | \Pi_i(\mathbf{x})}$ is fully equal to $\theta_{X_i = x_i | \Pi_i(\mathbf{x})}$.

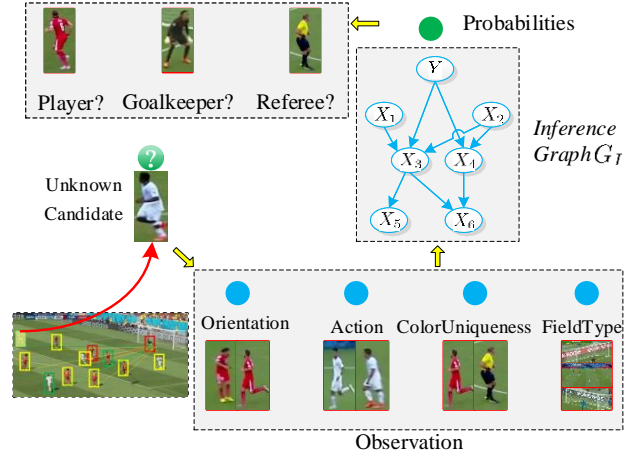

Figure 5: The pipeline of inference graph used for inferring the role of a person object.

With the notations above, the unique joint probability distribution of a Bayesian network (*i.e.*, the inference graph $G_i$) is given by

$$P_{\mathfrak{B}}(\mathbf{x}) = \prod_{i=1}^n \theta_{x_i | \Pi_i(\mathbf{x})} \qquad (1)$$

In our first problem, the purpose of Bayesian network is to infer the corresponding role that can be further regarded as an additional variable, *e.g.* $Y$ (similar handling for the second one). The notation of $Y$ is also a random variable associated with our target value with the values $y \in \mathcal{Y}$. In order to take $Y$ into consideration, we rearrange the data $\mathcal{D}$ into another form: $\mathcal{D} = \{(y^i, \mathbf{x}^{(i)})\}_{i=1}^N$. Accordingly, Eq. (1) is reformulated to the following form

$$P_{\mathfrak{B}}(y|\mathbf{x}) = \frac{P_{\mathfrak{B}}(y, \mathbf{x})}{P_{\mathfrak{B}}(\mathbf{x})} = \frac{\theta_{y|\Pi_i(\mathbf{x})} \prod_{i=1}^n \theta_{x_i|y, \Pi_i(\mathbf{x})}}{\sum_{y' \in \mathcal{Y}} \theta_{y'|\Pi_i(\mathbf{x})} \prod_{i=1}^n \theta_{x_i|y', \Pi_i(\mathbf{x})}} \qquad (2)$$

#### 4.4.2 **Learning the Inference Graph**

To preserve the significance of posterior estimator $P_{\mathfrak{B}}(y|\mathbf{x})$, Naïve Bayes takes the class variables as the root, and all attributes are conditional independent when conditioned on the class [23]. This assumption leads to the following form

$$P_{\mathfrak{B}}(y|\mathbf{x}) \propto \theta_y \prod_{i=1}^n \theta_{x_i|y} \qquad (3)$$

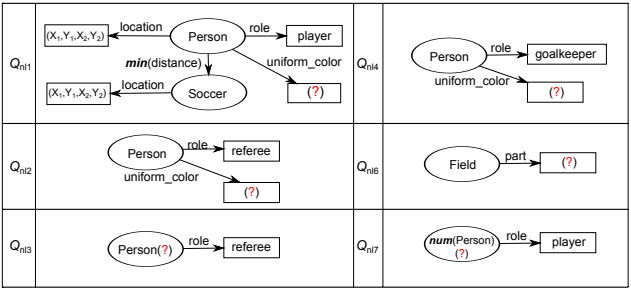
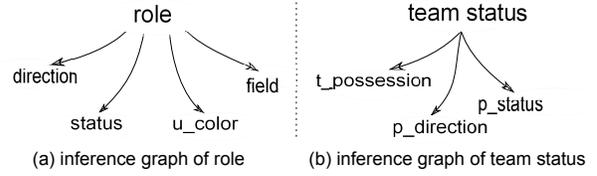

Figure 6: Query graphs

Figure 7: Inference Graphs

(a) inference graph of role    (b) inference graph of team status

As can be seen here, Naïve Bayes simplifies the structure of Bayesian network. In our proposed model, the structure of Naïve Bayes is used to infer the role of detected person.

To graphically and demonstratively infer the role of detected person, Figure 5 summarizes the pipeline of inference graph $G_I$, which are composed of two collaborative parts: state extraction (observation) and role probability inference. To be specific, orientation, action, color uniqueness of uniform, as well as field type are firstly employed to describe the state of an unknown candidate, which are then fed into the inference graph to produce the probability of each role. And the final role is decided based on the maximum probability.

After inference, one can either use a complete EAG to answer queries, or directly apply inference graph to find answers to certain queries (see Section 5 for an example).

### 4.5. EAG-based Matching

As introduced earlier, given a natural language query $Q_{nl}$, one needs to translate it into a query graph for evaluation. In light of this, we manually construct a set of query graphs, shown in Figure 6 as the correspondence of the set of questions given in Table 2. It is worth noting that query graph of $Q_{nl_5}$ is not provided as the query does not need matching computation.

One may notice that some pattern graphs are associated with functions on nodes or edges. The reason is that when transforming the questions into query graphs, we need to define some auxiliary functions to find correct answers.

Specifically, (1) we define the "min ()" function to measure the minimum distance, for the question "Who is holding the soccer?". The argument to min () is an array whose $i$-th element is the distance between the i-th player in the image and the soccer. Here the distance is Euclidean distance. (2) The "num ()" function is defined for the question "How many players are there in the image?". Its independent variable is all the *person* objects whose *role* attribute is "player" in the image, and the function value is the number of the independent variable.

Given an EAG that is generated from an image, we can answer queries as following. We first ignore functions de-fined on a query graph, and apply typical graph pattern matching algorithm, *e.g.* VF2 [5] to find matches. Over the set of matches of query graph, we operate arithmetic or set operations defined by functions, and obtain final answers.

## 5. Experiments

In this section, we conducted two sets of experiments to evaluate (1) the effectiveness of our inference module, and (2) the accuracy of our approach.

### 5.1. Effectiveness of Inference

To measure the performance of VI module, we define the inference accuracy following the F-measure [1]:

$$\mathsf{Acc}(A = \text{``}v\text{''}) = \frac{2 \cdot (\mathsf{recall}(A = \text{``}v\text{''}) \cdot \mathsf{precision}(A = \text{``}v\text{''}))}{(\mathsf{recall}(A = \text{``}v\text{''}) + \mathsf{precision}(A = \text{``}v\text{''}))},$$

where $\mathsf{recall}(A = \text{``}v\text{''}) = \frac{\#\text{true\_value\_inferred}}{\#\text{true\_value\_instance}}$, and $\mathsf{precision}(A = \text{``}v\text{''}) = \frac{\#\text{true\_value\_inferred}}{\#\text{inferred\_instance}}$. Here $\#\text{true\_value\_inferred}$ is the number of all the instances, whose attribute $A$ is inferred correctly as "$v$", $\#\text{true\_value\_instance}$ is the number of all the instances with attribute $A$ of value "$v$", and $\#\text{inferred\_instance}$ indicates the total number of instances whose attribute $A$ is inferred as "$v$".

| | $p(X|i = G)$ | $p(X|i = R)$ | $p(X|i = P)$ |
|---|---|---|---|
| *direction*="F" | 3.79 | 18.24 | 14.71 |
| *direction*="B" | 82.53 | 4.4 | 8.06 |
| *direction*="M" | 13.68 | 77.36 | 77.23 |
| *status*="E" | 47.59 | 0.47 | 4.46 |
| *status*="M" | 16.21 | 69.99 | 78.82 |
| *status*="S" | 34.02 | 27.36 | 14.3 |
| *status*="N" | 2.18 | 2.18 | 2.42 |
| *u_color*="M" | 4.02 | 20.89 | 99.36 |
| *u_color*="U" | 95.98 | 79.11 | 0.64 |
| *field*="L" | 51.38 | 16.76 | 15.01 |
| *field*="M" | 4.71 | 70.85 | 72.86 |
| *field*="R" | 43.91 | 12.39 | 12.13 |

Table 3: Conditional probability (%)

**Accuracy of Role**. Based on queries and image characteristics, we used four variables, *i.e. direction*, *status*, *field* and *unique_color* (abbr. *u_color*) to compute conditional probabilities. Figure 7(a) and Table 3 show inference graph and conditional probabilities, respectively. Note that the domain of variables *direction*, *status* and *field* are given in Table 1, while variable *u_color* can have one of two values, to indicate whether a *person* object has the unique uniform color (="U") or not (="M").

Using the conditional probabilities, VI infers *role* of each person object. The inference accuracy is shown in Table 4. One can find that the inference accuracies for different roles are above 85%, among which the accuracy even reaches 99% for role *player*.

|            | precision | recall | Acc  |
|------------|-----------|--------|------|
| role="*G*" | 94.4      | 85.5   | 89.8 |
| role="*R*" | 87.4      | 82.8   | 85   |
| role="*P*" | 98.8      | 99.3   | 99   |

Table 4: Inference accuracy of *role* (%). Here "*G*", "*R*" and "*P*" indicate goalkeeper, referee and player, respectively.

**Accuracy of Team Status**. Team status tells us whether a team is attacking or defending, it is closely related to question $Q_{nl_5}$. In practice, a defending team often has more players with "defending" status, and moreover, most of players are back to the goal. Based on this observation, we designed three variables, they are *p_status*, *p_direction* and *t_possession*, that represents players' status, players' direction and possession of the soccer, respectively. The domains of three variables are all $\{true, false\}$, where $true$ indicates that the team has more players with *expansion* status (resp. has more players back to the goal, has a player closest to the soccer), and $false$ otherwise.

Along the same line as computation of inference accuracy for *role*, we figure out inference accuracy for *team status*. Due to space constraint, we do not report conditional probabilities, but show inference accuracy and inference graph in Table 5 and Figure 7(b), respectively. As is shown, the inference accuracy reaches 81.3% when inferring whether a team is a defending team. Note that, in contrast to other questions, one can directly answer $Q_{nl_5}$ via inference, no matching computation is needed.

|                     | precision | recall | Acc  |
|---------------------|-----------|--------|------|
| Team Status="*D*"   | 89.8      | 74.2   | 81.3 |
| Team Status="*A*"   | 79.1      | 92.1   | 85.1 |

Table 5: Inference accuracy of *team status* (%). Here "*D*" and "*A*" indicate defending and attacking, respectively.

### 5.2. Overall Performance

We compared the following state-of-the-art methods: LSTM+CNN [4] and HieCoAttenVQA [19] with ours. As shown in Table 7 and 6, our approach is typically effective for *medium* and *hard* questions: (1) for *medium* questions, the average accuracy of our approach is 16.6% and 15.7% higher than that of LSTM+CNN and HieCoAttenVQA, respectively; and for *hard* questions, our approach substantially outperforms LSTM+CNN and HieCoAttenVQA, with average accuracy 3.59 and 3.56 times higher, than that of LSTM+CNN and HieCoAttenVQA, respectively. The advantage of our approach grows even larger for *hard* problems. (2) LSTM+CNN and HieCoAttenVQA work slightly better than our approach on simple questions, since they can easily learn correlations between images and questions, thus provide higher accuracy than ours. (3) Our method works best among three methods, as for all questions, the average accuracy of our approach is 38.1% and 31.9% higher than that of LSTM+CNN and HieCoAttenVQA, respectively.

|               | Easy  | Medium | Hard  | Average |
|---------------|-------|--------|-------|---------|
| LSTM+CNN      | 63.28 | 47.67  | 15.18 | 46.40   |
| HieCoAttenVQA | 68.26 | 48.04  | 15.26 | 49.11   |
| Ours          | **69.09** | **55.58** | **69.63** | **64.76** |

Table 6: Average accuracy comparison (%)

|            | LSTM+CNN | HieCoAttenVQA | Ours    |
|------------|----------|---------------|---------|
| $Q_{nl_1}$ | 44.23    | 43.62         | **71.58** |
| $Q_{nl_2}$ | 71.31    | **77.66**     | 64.88   |
| $Q_{nl_3}$ | 74.58    | **83.78**     | 70.8    |
| $Q_{nl_4}$ | 40.48    | 39.29         | **47.46** |
| $Q_{nl_5}$ | 49.19    | 49.90         | **63.7** |
| $Q_{nl_6}$ | 20.56    | 18.70         | **88.7** |
| $Q_{nl_7}$ | 11.08    | 12.63         | **50.55** |

Table 7: Accuracy comparison per query (%)

## 6. Conclusion

We propose a framework for understanding images regarding soccer matches and answering domain specific queries issued with natural languages. In contrast to previous works which learn correlation between images and answers, our method is able to do reasoning with inference graph $G_I$, and answer queries using structured query graph $Q$ and entity-attribute graph $G_{EA}$. Our idea on finding answers with graphs largely broaden the view in dealing with reasoning problems. Besides the approach, we also propose a new dataset, which is the first real-life dataset about soccer match in VQA literature. Experimental results show that our approach obtains better performance in accuracy, compared with the state of the art algorithms, furthermore, it significantly outperforms its counterparts for hard questions.

The study of graph-based VQA problem is still in its infancy. One issue is how to integrate external data, *e.g.* knowledge graph, for complicated reasoning tasks. Another issue concerns improvement of inference scheme, such that more hidden attributes can be inferred. The third topic is to design an interactive scheme for inference and visual tasks, thereby achieving better performances.